\documentclass{article}

\usepackage{geometry}

\usepackage{wrapfig}

\usepackage{enumitem}
\usepackage{natbib}
\setcitestyle{aysep={}} 
\usepackage{xcolor}
\usepackage{caption}

\usepackage{dsfont}
\usepackage{threeparttable}
\usepackage{amssymb}
\usepackage{mathtools}
\usepackage{float}
\usepackage{upgreek}
\allowdisplaybreaks
\usepackage{bm}
\usepackage{textcomp}
\usepackage{enumitem}
\usepackage{multirow}
\usepackage{colortbl}
\usepackage{makecell}
\usepackage{microtype}
\usepackage{graphicx}
\usepackage{subfigure}
\usepackage{booktabs}
\usepackage[linesnumbered,ruled,vlined]{algorithm2e}
\usepackage{amsthm}

\newcommand{\p}{p}

\newcommand{\e}{E}

\newcommand{\argmax}{\operatornamewithlimits{arg\max}}

\newcommand{\doo}{\textnormal{do}}
\allowdisplaybreaks

\newtheorem{remark}{Remark}

\newtheorem{lemma}{Lemma}

\usepackage{eucal}

\usepackage{pifont}
\newcommand{\cmark}{\ding{51}}  \newcommand{\xmark}{\ding{55}}  \usepackage{wrapfig}

\usepackage[utf8]{inputenc} \usepackage[T1]{fontenc}    \usepackage{hyperref}       \usepackage{url}            \usepackage{booktabs}       \usepackage{amsfonts}       \usepackage{nicefrac}       \usepackage{microtype}      \usepackage{xcolor}

\title{\bfseries Causal Policy Learning in Reinforcement Learning: Backdoor-Adjusted Soft Actor-Critic
}

\date{}

\author{
	Thanh Vinh Vo$^{1}$\quad Young Lee$^{ 1}$ \quad Haozhe Ma$^{ 1}$\quad Chien Lu$^{ 2}$\quad Tze-Yun Leong$^{ 1}$\\[0.2cm]
	$^{1}$National University of Singapore\\
	$^{2}$Trinity College Dublin, The University of Dublin
}

\begin{document}

	\maketitle
	
	\begin{abstract}

		\noindent Hidden confounders that influence both states and actions can bias policy learning in reinforcement learning (RL), leading to suboptimal or non-generalizable behavior. Most RL algorithms ignore this issue, learning policies from observational trajectories based solely on statistical associations rather than causal effects. We propose DoSAC (Do-Calculus Soft Actor-Critic with Backdoor Adjustment), a principled extension of the SAC algorithm that corrects for hidden confounding via causal intervention estimation. DoSAC estimates the interventional policy $\pi(a | \mathrm{do}(s))$ using the backdoor criterion, without requiring access to true confounders or causal labels. To achieve this, we introduce a learnable Backdoor Reconstructor that infers pseudo-past variables (previous state and action) from the current state to enable backdoor adjustment from observational data. This module is integrated into a soft actor-critic framework to compute both the interventional policy and its entropy. Empirical results on continuous control benchmarks show that DoSAC outperforms baselines under confounded settings, with improved robustness, generalization, and policy reliability.

	\end{abstract} 
	\section{Introduction}
	Reinforcement learning (RL) has enabled significant advancements in robotics, continuous control, and autonomous systems \citep{lillicrap2015ddpg, schulman2017proximal, haarnoja2018soft}. Among these, Soft Actor-Critic (SAC) \citep{haarnoja2018soft} stands out for its efficient exploration and strong empirical performance, making it particularly well-suited for continuous action spaces prevalent in real-world applications. However, like most RL algorithms, SAC relies primarily on observational data, implicitly assuming that observed state-action transitions accurately represent causal effects. This assumption can break down significantly when hidden confounders---unobserved variables simultaneously affecting the agent’s state observations and actions---are present. Such confounders introduce biases, impairing the reliability of learned value functions and policies, and thus limiting generalization and robustness \citep{bareinboim2015confounding, lu2021confounder}.
	
	To systematically address confounding, causal inference methods such as Pearl’s do-calculus \citep{pearl2009causality} have been developed to formalize and quantify causal relationships beyond mere associations. A key concept within this framework is the backdoor criterion, which provides a principled way to adjust for confounding biases by conditioning on appropriate variables that block spurious paths between cause and effect.
	
	Motivated by these ideas, we propose \textit{DoSAC} (Do-Calculus Soft Actor-Critic with Backdoor Adjustment), a causally-aware extension of SAC that explicitly estimates interventional policies to correct hidden confounding using the backdoor criterion. Instead of optimizing policies based on potentially biased observational distributions $\pi(a | s)$, DoSAC directly targets the interventional distribution $\pi(a | \mathrm{do}(s))$, thus capturing the genuine causal influence of states on actions. This shift facilitates more robust and generalizable decision-making, crucial for scenarios with subtle distributional shifts.
	
	Central to DoSAC is the \textit{Backdoor Reconstructor}, a novel neural module designed to infer pseudo-past variables (prior state and action) directly from the current state. These inferred variables serve as proxy adjustments to satisfy the backdoor criterion, enabling the estimation of interventional effects without requiring explicit supervision or access to latent confounders. Our main contributions are as follows:
	\begin{itemize}[noitemsep]
		\item[($1$)] We propose \textit{DoSAC}, a causally-informed extension of SAC designed to address hidden confounders by explicitly estimating interventional policies via backdoor adjustment, significantly improving robustness.
		
		\item[($2$)] We introduce the \textit{Backdoor Reconstructor}, a learnable neural component that effectively approximates the necessary conditioning set from the current state, which enables sampling from of the interventional policy using observational data from the replay buffer, and thus reducing confounding bias during training the AI agent.
		
		\item[($3$)] DoSAC integrates seamlessly into standard SAC training pipelines and can be trained efficiently end-to-end without additional overhead or data requirements.
		
		\item[($4$)] DoSAC generalizes SAC, naturally reducing to the original SAC formulation when no confounders affect action execution.
		
		\item[($5$)] Empirical evaluations demonstrate that DoSAC significantly improves policy robustness and generalization in continuous control tasks affected by synthetic confounding, outperforming standard SAC and a causality-inspired baseline.
	\end{itemize}

	\section{Background}
	
	\textbf{Reinforcement Learning (RL)} is commonly formulated as a Markov Decision Process (MDP), defined by the tuple \( \langle S, A, T, R, \gamma \rangle \), where \( \mathcal{S} \) is the state space, \( A \) is the action space, \( T: S \times A \times S \to [0,1] \) is the transition function, \( R: S \times A \times S \to \mathbb{R} \) is the reward function, and \( \gamma \in [0,1] \) is the discount factor. The agent follows a stochastic policy \( \pi: S \to \Delta(A) \) that maps states to distributions over actions, with the objective of maximizing the expected discounted return: $
	\mathbb{E}_{\tau} \left[ \sum_{t=0}^{\infty} \gamma^t R(s_t) \right]$, where a trajectory \( \tau = (s_0, a_0, s_1, a_1, \dots) \) is generated by sampling actions from the policy \( a_t \sim \pi(\cdot | s_t) \), and transitions from the environment \( s_{t+1} \sim T(\cdot | s_t, a_t) \). In model-free RL, popular approaches include value-based methods (e.g., Q-learning), policy gradient methods, and their hybrid, actor-critic algorithms~\citep{rl:sutton2018reinforcement}.

	\noindent \textbf{Causal Inference} studies cause-effect relationships, aiming to answer interventional questions such as ``What would happen if we changed variable \( X \)?'', especially important when learning from observational data confounded by some variables. Structural causal models (SCMs)~\citep{pearl2009causal} provide a formal framework for such reasoning using the do-calculus, which defines causal quantities like \( p(Y |\mathrm{do}(X)) \). An important tool is the \emph{backdoor adjustment}, which identifies the causal effect of \( X \) on \( Y \) by conditioning on a set of variables \( Z \) that block all backdoor paths: $
	p(Y | \mathrm{do}(X)) = \sum_Z p(Y | X, Z)\,p(Z)$. 
	In this work, we incorporate SCM principles into reinforcement learning by using backdoor adjustment to correct for hidden confounders that simultaneously influence both states and actions. This enables more robust and generalizable policy learning under observational bias.

	\section{Related Work}
	
	\textbf{Traditional RL.} 
	Conventional RL algorithms, such as Q-learning, policy gradients, and actor-critic methods, focus on learning policies that maximize expected cumulative rewards from observational trajectories \citep{lillicrap2015ddpg,schulman2017proximal,haarnoja2018soft}.  These methods generally assume fully observable and unbiased data, ignoring the presence of unobserved confounders. As a result, policies learned from these methods may exploit spurious correlations and generalize poorly when deployed in real-world systems where causal dependencies matter \citep{scholkopf2022causality,zhang2020causal}.

	\noindent\textbf{Causal RL.} Prior work in causal reinforcement learning spans diverse goals, including representation learning, policy learning, and off-policy evaluation. Some methods focus on learning invariant policies across changing environments~\citep{namkoong2020off,zhang2020designing,zhang2019near,zhang2021bounding}, while others impose structural constraints on the policy space using causal graphs~\citep{lee2019structural,lee2020characterizing,zhang2022online}. Causal imitation learning has also gained interest~\citep{kumor2021sequential,ruan2023causal,swamy2022causal,zhang2020causal}, as well as work on counterfactual reasoning~\citep{bareinboim2015bandits,forney2019counterfactual,zhang2016markov,zhang2022can} and causal model-based action effect estimation~\citep{ghassami2018budgeted,jaber2020causal}.
	
	Several works address hidden confounding in RL. \citet{oberst2019counterfactual} propose using structural causal models (SCMs) for off-policy evaluation. \citet{cai2024learning} introduce an online framework that learns both the causal graph and the policy to correct for confounders. DOVI~\citep{wang2021provably} provides a theoretically grounded method for causal policy learning via backdoor adjustment, though it lacks empirical evaluation and code. Delphic RL~\citep{pace2024delphic} and Two-way Deconfounder~\citep{yu2024two} target offline policy evaluation under unobserved confounding, while Shi et al.~\citep{shi2022minimax} address confounded off-policy evaluation in POMDPs. Other works focus on causal representation learning through exploration~\citep{sontakke2021causal} or adjust sample influence via causal signals~\citep{zhu2023causal}.
	
	The most closely related method to ours is ACE~\citep{ji2024ace}, which operates in an online, off-policy setting and incorporates causality-aware entropy regularization to guide exploration. However, ACE assumes structured action decompositions and does not address hidden confounding. In contrast, our method DoSAC explicitly targets action-level confounding by estimating the interventional policy \( \pi(a | \mathrm{do}(s)) \) via backdoor adjustment. We introduce a neural Backdoor Reconstructor to infer adjustment variables from the current state, enabling causal policy learning without counterfactual supervision or structural priors. A summary of the key differences between DoSAC and existing causal RL methods is provided in the  Appendix.

	\section{DoSAC: Do-Calculus Soft Actor-Critic with Backdoor Adjustment}
	In this section, we first describe the underlying structural causal model and define the interventional policy objective. We then introduce causal entropy as a regularizer and propose a backdoor adjustment mechanism with a learnable reconstructor. Finally, we present the full learning algorithm and provide theoretical results showing that DoSAC generalizes and reduces to SAC in the absence of confounding.

	\subsection{The Model}
	\label{sec:the-model}
\begin{figure}
		\centering
		\includegraphics[width=1.0\linewidth]{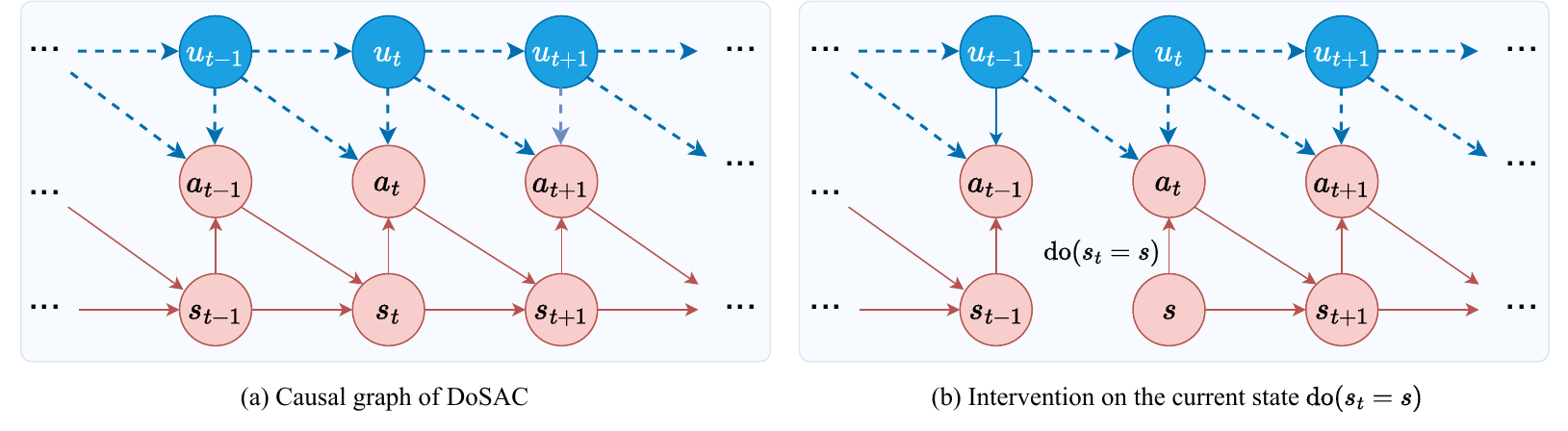}
		\caption{
			(a) Structural causal model illustrating hidden confounders \( u_t \) that affect both past and current actions, inducing spurious dependencies between state \( s_t \) and action \( a_t \). Standard RL methods learn policies based on \( \pi(a_t | s_t) \), which may be biased due to unblocked backdoor paths. 
			(b) In contrast, DoSAC targets the interventional distribution \( \pi(a_t | \mathrm{do}(s_t)) \), which blocks all backdoor paths by intervening on the current state, thereby removing confounding bias and enabling robust policy learning.
			\label{fig:model}
		}
	\end{figure}

	Consider an RL problem $\langle S, A, T, R, U, \gamma \rangle$, where $S$, $A$, $T$, $R$, and $\gamma$ are the state space, action space, transition function, reward function, and discount factor. The additional component $U$ represents the space of stochastic confounder of actions and this is a latent space. We illustrate the proposed model using causal graph in Figure~\ref{fig:model}(a). The figure shows that the traditional policy would be learned by combination of the causal relationship $s_t \rightarrow a_t$ and the `backdoor' paths $s_t \leftarrow a_{t-1} \leftarrow u_{t-1} \rightarrow a_t$, and $s_t \leftarrow a_{t-1} \leftarrow u_{t-1} \rightarrow u_t \rightarrow a_t$. A typical RL algorithm, which learns the policy $\pi(a|s)$, would ignore confounders and hence results in a biased model. Such a model would not be applicable to a new scenario. For example, when the effect of the confounders $u_t$ is different from the training data or when there are no confounders. 
	
	In this work, we aim to learn a robust policy which is invariant to the confounder bias. We propose learning the interventional policy $\pi(a|\doo(s))$, which decides the next action based on intervention on the current state. We illustrate this intervention in Figure~\ref{fig:model}(b). Once an intervention $\doo(s_t=s)$ is performed, the bias effects of the confounders are eliminated, i.e., there is no backdoor path under an intervention. Note that performing an intervention $\doo(s)$ is intractable as we cannot set the state of the agent to a specific state. Hence, we need to learn it from the observational data from the replay buffer, which contains spurious correlations induced by the confounders $u_t$.
	
	\subsection{Causal Entropy}
	We propose causal entropy which measures the amount of disorder or randomness of the action with respect to an interventional policy. It is defined as
	\begin{align}
		H(a|do(s)) = \e_{a\sim \pi}\big[-\log\pi(a|\doo(s))\big],
	\end{align}
	where $\doo(s)$ represents an intervention on the state by setting it to $s$, and $\pi(a|\doo(s))$ is the causal policy. It is clear that the causal policy and entropy are different from the traditional policy and entropy in that they are conditioned on the intervention of the state, i.e., $\pi(a|\doo(s)) \neq \pi(a|s)$ and $H(a|\doo(s)) \neq H(a|s)$. 
	
	By maximizing causal entropy, the agent seeks to make its policy as ``random'' or uncertain as possible with respect to the causal policy, and hence learning a robust policy which is invariant to the confounders.
	
	\begin{remark}
		The causal entropy can recover the traditional entropy once there are no confounders, hence offering a more robust and general model.
	\end{remark}
	
	In this work, we incorporate causal entropy into the MDP objective as an auxiliary reward to promote exploration and improve robustness to confounding.
	
	\subsection{The Learning Algorithm}
	
	We optimize a bonus reward at each time step proportional to the causal entropy policy at that time step:
	\begin{align}
		\pi^* \!\!=\! \argmax_{\pi}\!\underset{\tau\sim\pi}{\e\!}\!\left[\sum_{t=0}^{\infty}\gamma^t\Big(R(\cdot) \!+\! \alpha H(\cdot|\doo(s_t))\Big)\right]\!\!,\label{eq:pi}
	\end{align}
	where $R(\cdot)=R(s_t, a_t, s_{t+1})$ is the reward, and the expectation is taken over trajectories drawn from the inerventional policy $ \pi(a|\doo(s))$. Eq~(\ref{eq:pi}) is adapted from the SAC algorithm \citep{haarnoja2018soft}. Herein, we included the causal entropy and policy to learn a robust and general policy since it would remove bias effects from the confounders of the actions.

	\noindent\textbf{Backdoor Reconstructor and Interventional Actor.} 
	From Eq.~(\ref{eq:pi}), drawing actions from the interventional policy $\pi(a_t | \mathrm{do}(s_t))$ would require intervening or randomizing the current state, which is often infeasible in practice. To address this, we invoke the backdoor adjustment formula to express the interventional policy using observational data:
	\begin{align}
		\pi(a_t | \mathrm{do}(s_t=s)) 
		= \mathbb{E}_{(a_{t-1}, s_{t-1}) \sim p(a_{t-1}, s_{t-1})} 
		\left[ p(a_t | s_t=s, a_{t-1}, s_{t-1}) \right], \label{eq:backdoor}
	\end{align}
	where both terms on the right-hand side can be estimated from samples in the replay buffer. The conditional distribution $p(a_t | s_t, a_{t-1}, s_{t-1})$ can be learned directly from observed tuples $(a_t, s_t, a_{t-1}, s_{t-1})$ collected during training. 
	
	To approximate the marginal distribution $p(a_{t-1}, s_{t-1})$, we leverage a model-based approach: we infer pseudo-past variables from the current state. Specifically, we define a dummy variable $\tilde{s}_t$ representing the current state, and write:
	\begin{align}
		p(a_{t-1}, s_{t-1}) = \int p(a_{t-1}, s_{t-1} | \tilde{s}_t) \, p(\tilde{s}_t) \, d\tilde{s}_t.
	\end{align}
	In practice, we approximate this by setting $\tilde{s}_t = s_t$ using states sampled from the replay buffer, and learn a conditional model $p_\phi(a_{t-1}, s_{t-1} | s_t)$ to predict the pseudo-past from the current state.

	Putting these components together, we parameterize the interventional policy $\pi_\theta(a_t | \mathrm{do}(s_t))$ using two learned conditional distributions:
	\begin{align}
		p_\theta(a_t | s_t, a_{t-1}, s_{t-1}) \quad \text{and} \quad p_\phi(a_{t-1}, s_{t-1} | s_t). \label{eq:cond-dists}
	\end{align}
	To sample an action from the interventional policy, we use forward sampling: we first draw $(a_{t-1}, s_{t-1}) \sim p_\phi(\cdot | s_t)$, then sample $a_t \sim p_\theta(\cdot | s_t, a_{t-1}, s_{t-1})$. The procedure is summarized in Algorithm~\ref{alg:sampling-pi}.
	
	{\IncMargin{1.2em}
		\begin{algorithm}\caption{Sampling from the causal policy $\pi(\cdot|\doo(s))$}
			\label{alg:sampling-pi}
			
			\SetKwInOut{Input}{Input}
			\Input{\, The state $s$.}
			\Begin{
				Draw a sample $(\tilde{a}, \tilde{s}) \sim q_\phi(a_{t-1},s_{t-1}|s_t=s)$;
				
				Draw a sample $a \sim p_\theta(a_{t}|s_t=s, a_{t-1}=\tilde{a}, s_{t-1}=\tilde{s})$;
				
				\textbf{Return} $a$;
			}
		\end{algorithm}
		\DecMargin{1.2em}
	}
	
	\begin{figure}[t]
		\centering
		\includegraphics[width=0.85\linewidth]{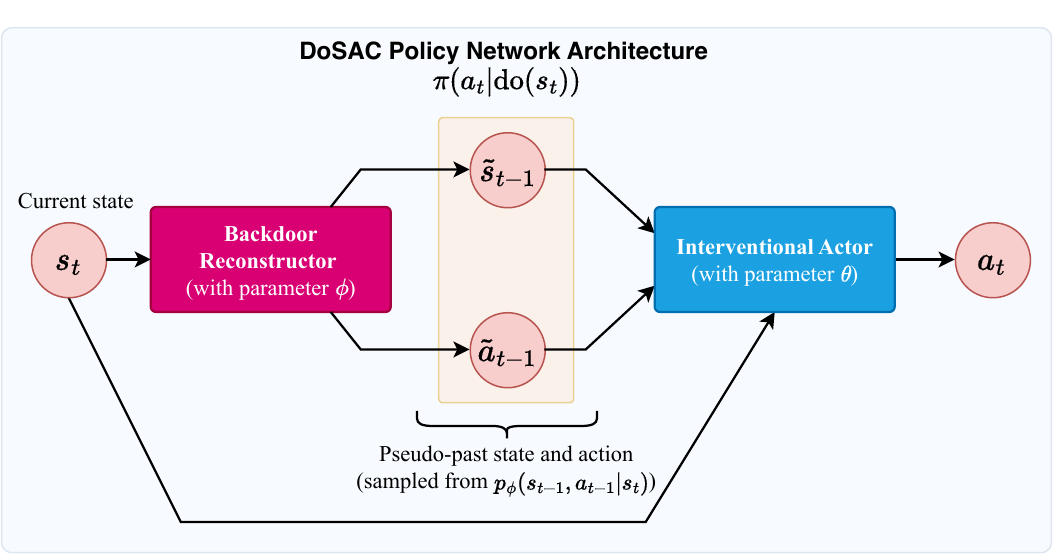}
		\caption{
			DoSAC Policy Network Architecture: The Backdoor Reconstructor (parameterized by \( \phi \)) infers pseudo-past variables \( (\tilde{s}_{t-1}, \tilde{a}_{t-1}) \sim p_\phi(s_{t-1}, a_{t-1} | s_t) \), which are then used by the Interventional Actor (parameterized by \( \theta \)) along with \( s_t \) to define the distribution \( \pi(a_t | \mathrm{do}(s_t)) \). This architecture enables backdoor-adjusted action sampling using observational data alone.}
		\label{fig:dosac-architecture}
		\vspace{-12pt}
	\end{figure}
	
	Figure~\ref{fig:dosac-architecture} illustrates the architecture of the DoSAC policy network. Starting from the current state \( s_t \), the \textit{Backdoor Reconstructor}, a neural network trained to approximate the inverse dynamics, predicts pseudo-past variables \( (\tilde{s}_{t-1}, \tilde{a}_{t-1}) \), which are treated as surrogates for the true but unobserved past context. These pseudo-past variables are then used by the \textit{Interventional Actor} to sample actions from the adjusted distribution \( \pi(a_t | \mathrm{do}(s_t)) \), in accordance with the backdoor criterion. This two-stage mechanism enables the policy to mitigate bias from hidden confounders by approximating causal interventions using only observational data. In the following, we present the end-to-end training procedure.
	
	\noindent\textbf{Training the model.} 
	The value function $V^\pi(s)$ and Q-function $Q^\pi(s,a)$ would include the interventional entropy  from every timestep.  We have the following relationship:
	\begin{align}
		&V^\pi(s) = \underset{a\sim\pi}{\e}\left[Q^\pi(s,a)\right] + \alpha H(\cdot|\doo(s)),\label{eq:v-recursive}\\
		&Q^\pi(s,a) = \underset{\substack{s'\sim P\\ a'\sim\pi}}{\e}\left[R(\cdot) + \gamma(Q^\pi(s',a') + \alpha H(\cdot|\doo(s')))\right],\label{eq:q-recursive}
	\end{align}
	where $s, a, s', a'$ are the current state, current action, next state, and next action, respectively. The right hand side in Eq.~(\ref{eq:q-recursive}) can be approximated by drawing samples from the replay buffer and the current interventional policy, i.e.,
	\begin{align}
		Q^\pi(s,a) \simeq r + \gamma(Q^\pi(s', \tilde{a}') + \alpha \log \p(\tilde{a}'|s', \tilde{s},\tilde{a})),
	\end{align}
	where $\tilde{s},\tilde{a} \sim \p(a, s|s')$ and $\tilde{a}' \sim \p(\tilde{a}'|s', \tilde{s},\tilde{a})$ .
	
	We can learn approximation of the Q-function by parameterize it and minimizing the left and right hand side of Eq.~(\ref{eq:q-recursive}). In particular, we minimize the loss function:
	\begin{align}
		L = \e_{s,a,r,s',d\sim D}\Big(Q_\psi(s,a) - f(r,s',d)\Big)^2,
	\end{align}
	where the target $f(r,s',d)$ is calculated as follows: $f(r,s',d) = r + \gamma(1-d)[Q_\psi(s',a') - \alpha \log \pi_{\theta,\phi}(a'|\doo(s'))]$, where $a'$ is sampled from $\pi_{\theta,\phi}(a'|\doo(s'))$ using Algorithm~\ref{alg:sampling-pi}, and $\theta, \phi, \psi$ are sets of parameters to be optimized.

	To provide theoretical grounding for our method, we establish Lemma~\ref{lem:causal-sac} as follows:
	\begin{lemma}\label{lem:causal-sac}
		Let $Q^\pi(s, a)$ be the soft Q-function defined under the interventional distribution $\mathrm{do}(s)$, and define the soft policy improvement as 
		$\pi_{\mathrm{new}}(a \mid \mathrm{do}(s)) \propto \exp\left( \frac{1}{\alpha} Q^\pi(s, a) \right)$. 
		Then the entropy-regularized causal objective
		\[
		J(\pi) = \mathbb{E}_{s \sim p(s)} \left[ \mathbb{E}_{a \sim \pi(a | \mathrm{do}(s))} \left[ Q^\pi(s, a) - \alpha \log \pi(a | \mathrm{do}(s)) \right] \right]
		\]
		satisfies $J(\pi_{\mathrm{new}}) \ge J(\pi)$, 
		with equality if and only if $\pi_{\mathrm{new}} = \pi$.
	\end{lemma}
	Please refer to the Appendix for the proof of Lemma~\ref{lem:causal-sac}. Lemma~\ref{lem:causal-sac} guarantees that alternating between policy evaluation and soft policy improvement under the interventional distribution leads to monotonic improvement in the entropy-regularized causal objective. Specifically, by updating the policy toward a Boltzmann distribution over the causal Q-values--estimated using backdoor-adjusted trajectories--we ensure that each iteration either improves or maintains performance with respect to the interventional reward. This result mirrors classical policy improvement guarantees in standard RL but is derived under the causal setting, where hidden confounders may bias observed trajectories. Our theorem highlights that, despite relying only on observational data, DoSAC performs principled updates that reflect the true causal effect of actions, enabling stable learning in confounded environments.

	\begin{lemma}[Reduction to Standard SAC]
		\label{lem:reduction-sac}
		Assume the environment has no hidden confounders $u_t$s. Then the interventional distribution $\pi(a_t | \mathrm{do}(s_t))$ is equal to the observational distribution $\pi(a_t | s_t)$, and the DoSAC objective reduces exactly to the standard SAC objective.
	\end{lemma}

	Lemma~\ref{lem:reduction-sac} confirms that DoSAC is a strict generalization of SAC: when no confounding exists, the interventional distribution $\pi(a_t | \mathrm{do}(s_t))$ collapses to the standard observational policy $\pi(a_t | s_t)$, and all components of DoSAC--policy evaluation, policy improvement, and training dynamics--reduce to those in SAC. In this setting, the Backdoor Reconstructor becomes functionally redundant, as the model $p_\phi(a_{t-1}, s_{t-1} | s_t)$ is no longer required to block confounding paths. This guarantees that DoSAC introduces no unnecessary overhead or divergence from standard RL behavior when the environment does not contain confounders.

	\section{Experiment}
	
	\label{sec:experiment}

	\textbf{Baselines.} We first compare DoSAC directly with SAC to demonstrate the improvements introduced by our method. Subsequently, we benchmark DoSAC against a strong suite of baselines that ($i$) support online reinforcement learning, ($ii$) learn explicit agent policies, and ($iii$) have publicly available implementations. These criteria ensure reproducibility and fair comparisons under consistent training protocols. Specifically, we include SAC (Soft Actor-Critic)\citep{haarnoja2018soft} as a standard off-policy baseline, ACE (Causality-Aware Entropy Regularization)\citep{ji2024ace} as a recent causality-aware method operating in the online setting, as well as widely used RL methods such as TD3 (Twin Delayed Deep Deterministic Policy Gradient) \citep{fujimoto2018addressing}, PPO (Proximal Policy Optimization) \citep{schulman2017proximal}, DDPG (Deep Deterministic Policy Gradient) \citep{lillicrap2015continuous}, and RPO (Robust Policy Optimization)~\citep{rahman2022robust}. Methods lacking publicly available code or not involving explicit policy learning are excluded to maintain the fairness and reproducibility of our experiments.
	Implementation of the baselines SAC, PPO, TD3, DDPG, RPO are based on the \texttt{CleanRL} library \citep{huang2022cleanrl}. Implementation of ACE is taken from \citep{ji2024ace}. 
	
	The experimental evaluation is on standard continuous control benchmarks from the OpenAI Gym suite \citep{brockman2016openai}, including \texttt{Humanroid}, \texttt{Ant}, \texttt{Walker2d}, and \texttt{LunarLander}. To ensure fair comparisons, all methods are trained using the same network architecture, replay buffer size, batch size, and number of environment steps.
	
	\noindent\textbf{Injecting confounding bias.}
	To simulate hidden confounding in action selection, we modify the agent’s actions during both training and evaluation by introducing an additive confounder \( u_t \sim N(\mu, \sigma^2 I) \), where \( \mu \) and \( \sigma \) are task-specific constants and \( I \) is the identity matrix matching the action dimension. Specifically, at each timestep, the agent samples its nominal action \( a_t \sim \pi(a_t | s_t) \), and the environment receives a confounded action \( \tilde{a}_t = a_t + u_t \). This simulates the presence of a hidden confounder that influences both the observed state (via the environment dynamics) and the chosen action. This setup creates spurious correlations in the observational data, allowing us to evaluate whether DoSAC can recover interventional policies through backdoor adjustment. During evaluation, we test the agent both under the same confounding distribution and under clean conditions (\( u_t = 0 \)) to assess robustness and generalization.

	\begin{figure}
		\centering
		\includegraphics[width=0.99\linewidth]{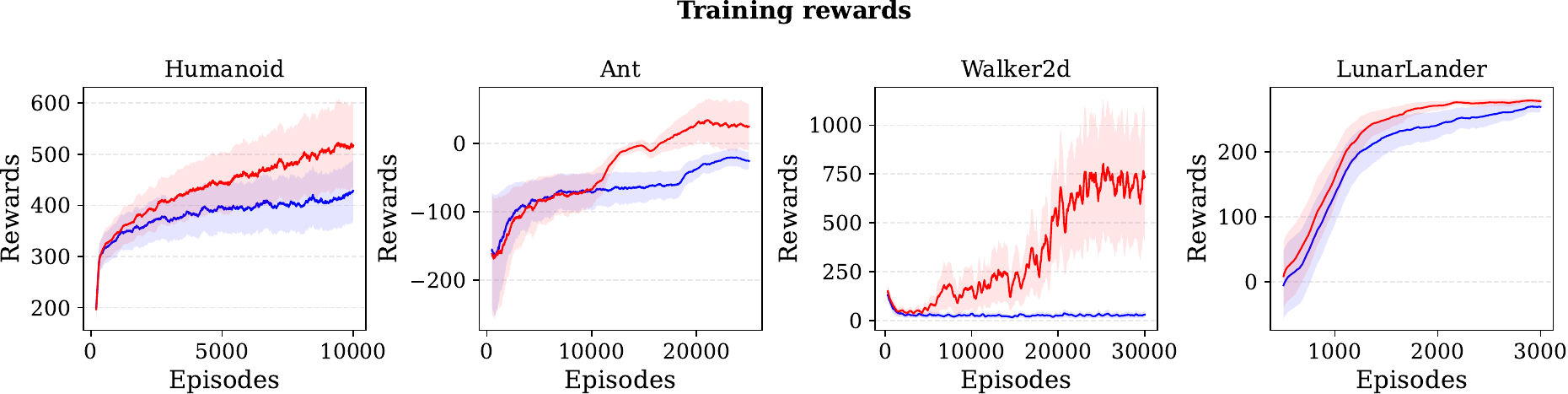}
		\\[0.3cm]
		\includegraphics[width=0.99\linewidth]{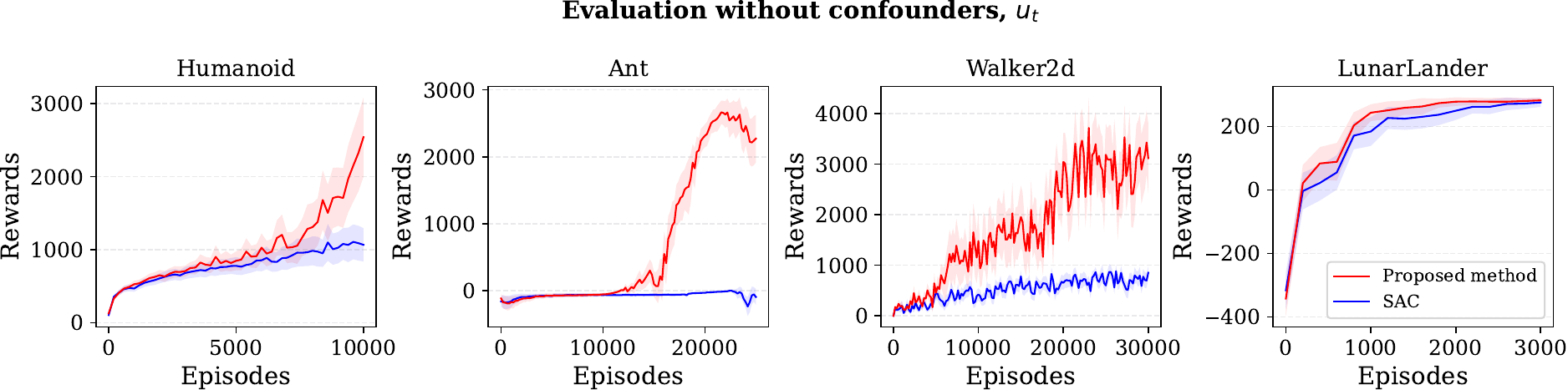}
		\\[0.2cm]
		\includegraphics[width=0.99\linewidth]{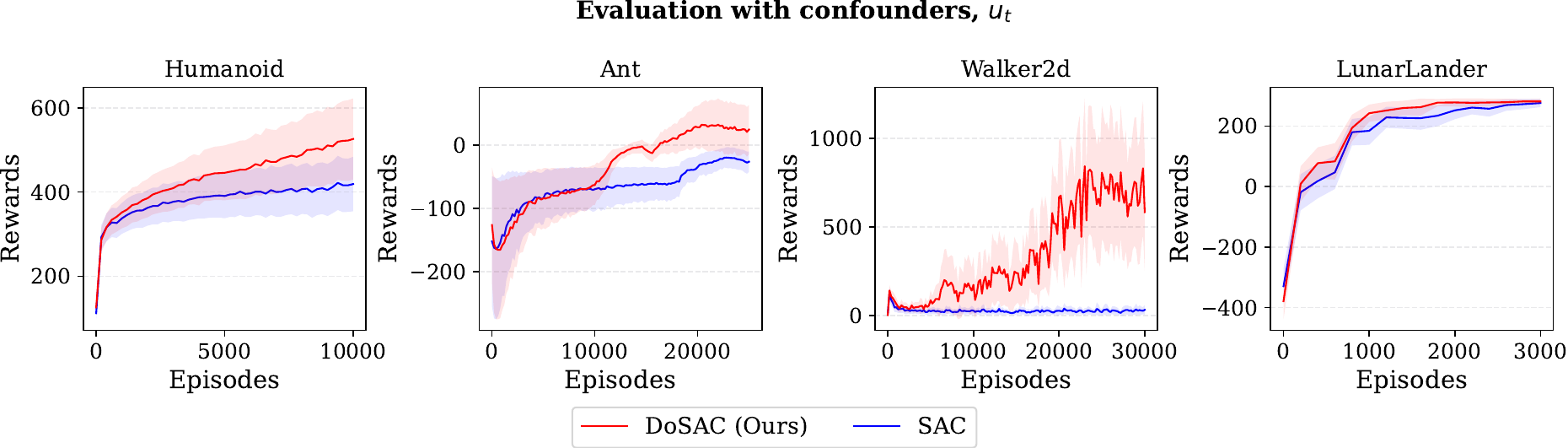}
		\caption{Training and evaluation performance across four environments. \textit{Top row}: Training rewards over episodes. \textit{Middle row}: Evaluation rewards without confounders. \textit{Bottom row}: Evaluation rewards with confounders. Across all settings, the proposed method outperforms SAC, with notable improvements in high-dimensional environments. It generalizes better to unseen test scenarios and shows increased robustness to confounding variables.}
		\label{fig:DoSAC-vs-SAC}
		\vspace{-12pt}
	\end{figure}

	\subsection{DoSAC vs. SAC: An Ablation Study} 
	Before comparing with other baselines, we first conduct an ablation study between DoSAC and SAC to isolate the impact of our proposed modifications, as DoSAC builds directly upon the SAC algorithm. Figure~\ref{fig:DoSAC-vs-SAC} summarizes the training and evaluation performance. During training, DoSAC consistently achieves higher cumulative rewards, particularly evident in complex, high-dimensional tasks such as \textit{Ant} and \textit{Walker2d}, demonstrating enhanced sample efficiency and stable policy learning compared to SAC. Remarkably, even without confounders at evaluation, DoSAC significantly outperforms SAC, clearly highlighting its superior generalization capabilities driven by explicit causal modeling. Under evaluation with confounders, both methods experience performance degradation; however, DoSAC remains substantially more robust, notably maintaining stable performance in tasks like \textit{Ant} and \textit{Walker2d}, whereas SAC experiences severe sensitivity to confounding shifts. These results underscore the practical strength of explicitly addressing confounding in policy learning, validating the theoretical motivations for our causal approach, and highlighting its advantages in robustness, generalization, and reliability in off-policy reinforcement learning.
	
	\subsection{Compare with the Baselines.} 
	
	In this section, we compare DoSAC with the baselines. For all the methods, we set a global time step to 2,000,000 for \texttt{Ant}, \texttt{Walker2d}, and \texttt{Humanoid}, and 1,000,000 steps for \texttt{LunarLander}. For each methods, we perform 5 runs with different random seeds and report the average and standard error of the expected returns. After training, we evaluate the learned agent on two cases: \emph{with confounders} and \emph{without confounders}. We also study their performance on different noisy strengths of the latent confounder $u_t$ to study their sensitivity to the noisy confounders.  
	
	\noindent\textbf{Evaluation without Confounders.} In Table~\ref{tab:eval-without-confounders}, agents trained with hidden confounders are evaluated in clean (unconfounded) environments, highlighting the generalization capability of each method. DoSAC consistently achieves significantly higher returns across all environments, surpassing other methods by a substantial margin. Particularly noteworthy are the performance improvements in complex environments such as Ant and Humanoid, where DoSAC attains returns of approximately 2252 and 2361, respectively, greatly outperforming standard SAC (1553 and 1079) and ACE (1400 and 1217). In Walker2d, DoSAC demonstrates nearly double the performance of SAC and markedly outperforms ACE. This clearly indicates the robustness and superior generalization capability of DoSAC when adapting from confounded training scenarios to clean test conditions.
	
	\noindent\textbf{Evaluation with Confounders.} Table~\ref{tab:eval-with-confounders} evaluates methods under persistent confounding during testing, reflecting robustness under ongoing confounded conditions. Here, while performance drops are observed across all methods compared to the unconfounded test scenario, DoSAC maintains substantial advantages in all tasks. Notably, in Ant and Humanoid, DoSAC is the only method that achieves significant positive returns (45.4 and 524.5 respectively), while other methods suffer substantial degradation, frequently producing negative or negligible returns. Similarly, in Walker2d and LunarLander, DoSAC consistently outperforms SAC, ACE, and other baselines, further underscoring its resilience to confounding influences.
	
	Overall, the experimental evidence strongly supports the effectiveness of the DoSAC method in improving both generalization and robustness to hidden confounding, making it particularly suitable for real-world applications where confounding factors are prevalent.

	\begin{table}
		\centering
		\caption{Evaluation in environments \emph{without confounders}. All agents were trained in the presence of hidden confounders, but evaluated in clean environments. We report the average return ($\pm$ standard error) over 5 runs for each method across four continuous control tasks. This setting highlights the ability of each method to generalize beyond confounded training conditions.}
		\vspace{3pt}
		\label{tab:eval-without-confounders}
		\small
		\setlength{\tabcolsep}{4.3pt}
		\begin{tabular}{lcccc}
			\toprule
			Method & \texttt{Ant} & \texttt{Humanoid} & \texttt{Walker2d} & \texttt{LunarLander} \\
			\midrule
			PPO \citep{schulman2017proximal}     & $-1.4 \pm 0.8$ & $5.0 \pm 0.1$ & $-0.1 \pm 1.8$ & $-8.7 \pm 0.6$ \\
			RPO \citep{rahman2022robust}     & $-164.8 \pm 67.3$ & $4.9 \pm 0.1$ & $4.6 \pm 0.1$ & $-4.9 \pm 1.7$ \\
			DDPG \citep{lillicrap2015continuous}    & $826.1 \pm 91.4$ & $523.7 \pm 40.1$ & $191.8 \pm 33.8$ & $104.3 \pm 15.0$ \\
			TD3 \citep{fujimoto2018addressing}     & $765.9 \pm 33.0$ & $857.0 \pm 77.6$ & $2437.6 \pm 171.8$ & $273.1 \pm 1.6$ \\
			SAC \citep{haarnoja2018soft}     & $1553.0 \pm 162.7$ & $1078.9 \pm 73.0$ & $651.6\pm 56.9$ & $271.1 \pm 5.9$ \\
			ACE \citep{ji2024ace}    & $1400.4 \pm 140.5$ & $1217.5 \pm 91.2$ & $3572.7 \pm 215.5$ & $73.9 \pm 34.2$ \\\midrule
			DoSAC (Ours) & $\mathbf{2252.3 \pm 103.7}$ & $\mathbf{2361.1 \pm 127.0}$ & $\mathbf{3983.5 \pm 173.1}$ & $\mathbf{282.2 \pm 1.0}$ \\
			\bottomrule
		\end{tabular}
		\vspace{-16pt}
	\end{table}

	\begin{table}
		\centering
		\caption{Evaluation in environments \emph{with confounders}. All agents were trained in the presence of hidden confounders, but evaluated in clean environments. We report the average return ($\pm$ standard error) over 5 runs for each method across four continuous control tasks. This setting highlights the ability of each method to generalize beyond confounded training conditions.}
		\vspace{3pt}
		\label{tab:eval-with-confounders}
		\small
		\setlength{\tabcolsep}{7pt}
		\begin{tabular}{lcccc}
			\toprule
			Method & \texttt{Ant} & \texttt{Humanoid} & \texttt{Walker2d} & \texttt{LunarLander} \\
			\midrule
			PPO \citep{schulman2017proximal}     & $-2.7 \pm 0.5$ & $4.5 \pm 0.1$ & $0.9 \pm 2.1$ & $-7.4 \pm 0.6$ \\
			RPO \citep{rahman2022robust}     & $-268.0 \pm 119.3$ & $4.6 \pm 0.1$ & $4.6 \pm 0.1$ & $-3.4 \pm 1.8$ \\
			DDPG \citep{lillicrap2015continuous}   & $-101.0 \pm 9.4$ & $342.5 \pm 10.9$ & $112.1 \pm 31.6$ & $71.5 \pm 27.2$ \\
			TD3 \citep{fujimoto2018addressing}     & $-163.5 \pm 8.8$ & $402.3 \pm 17.5$ & $386.5 \pm 26.1$ & $281.7 \pm 1.8$ \\
			SAC \citep{haarnoja2018soft}    & $39.2 \pm 6.7$ & $420.9 \pm 15.9$ & $31.1 \pm 7.9$ & $271.2 \pm 5.7$ \\
			ACE  \citep{ji2024ace}   & $14.4 \pm 5.1$ & $430.7 \pm 19.3$ & $813.3 \pm 88.1$ & $-21.6 \pm 8.0$ \\
			\textbf{DoSAC (Ours)} & $\mathbf{45.4 \pm 5.8}$ & $\mathbf{524.5 \pm 20.1}$ & $\mathbf{820.9 \pm 73.1}$ & $\mathbf{282.5 \pm 1.0}$ \\
			\bottomrule
		\end{tabular}
		\vspace{-6pt}
	\end{table}
	
	\begin{figure}
		\centering
		\includegraphics[width=0.99\linewidth]{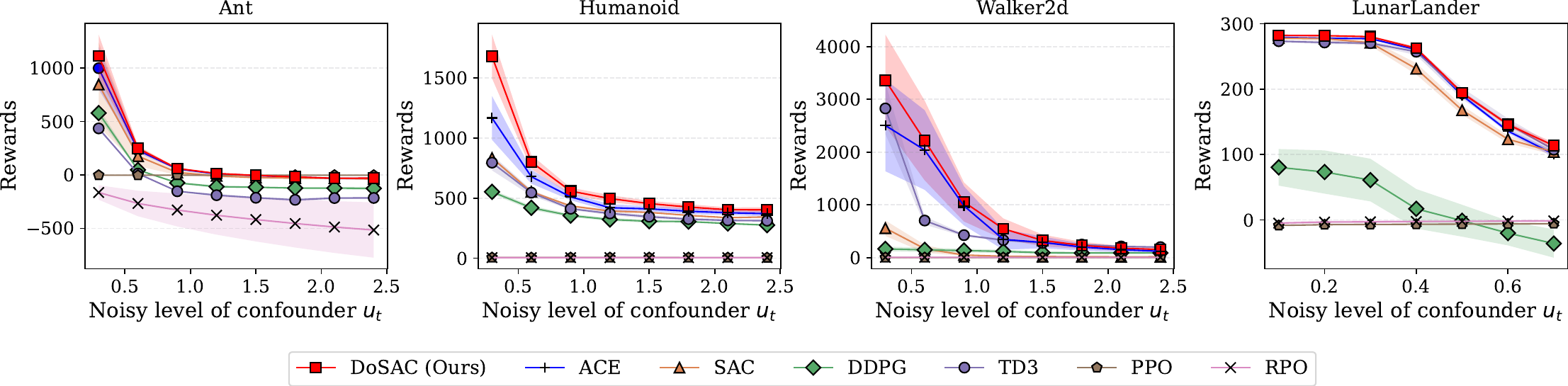}
		\caption{The expected rewards accross different strength level of the confounders $u_t$.}
		\label{fig:compare-baselines}
		\vspace{-12pt}
	\end{figure}
	
	\noindent\textbf{Sensitivity to Confounder Strength.} In this experiment, we evaluate with different noisy strengths of the confounders $u_t$. In particular, we set different values for the standard deviations $\sigma$ of $u_t$, $\sigma \in \{0.3, 0.6, 0.9, 1.2, 1.5, 1.8, 2.1, 2.4\}$ for \texttt{Ant}, \texttt{Humanoid}, and \texttt{Walker2d}, and  $\sigma \in \{0.1, 0.2, 0.3, 1.4, 0.5, 0.6, 0.7\}$ for \texttt{LunarLander}. Figure~\ref{fig:compare-baselines} illustrates the performance of different reinforcement learning methods across varying strengths of confounders.  DoSAC consistently demonstrates superior robustness, achieving higher returns compared to other baseline methods. As the confounder strength increases, performance degradation is observed universally across methods; however, DoSAC exhibits significantly less sensitivity to rising confounder levels. In particular, in complex tasks like Ant, Humanoid, and Walker2d, DoSAC maintains notably higher performance levels even under strong confounding conditions, highlighting its capability to effectively mitigate the influence of hidden confounders. Other methods, including SAC and ACE, suffer more substantial performance losses, underscoring the importance of explicit causal adjustment implemented in DoSAC.

	\section{Conclusion}

	We introduced DoSAC, a backdoor-adjusted extension of Soft Actor-Critic, leveraging causal inference to effectively address hidden confounders in off-policy reinforcement learning. By reconstructing pseudo-past variables from current states and applying the backdoor criterion, DoSAC estimates interventional policies from purely observational data. This approach integrates seamlessly into actor-critic architectures and is supported by theoretical analyses and empirical evidence, demonstrating substantial improvements in robustness, generalization, and overall performance across various continuous control tasks. 
	
	The ability of DoSAC to effectively handle hidden confounding can significantly benefit real-world applications where observational data often contain unobserved biases, such as healthcare decision-making, autonomous driving, and robotics. By learning more reliable, causally robust policies, DoSAC has the potential to enhance the safety, interpretability, and generalization of reinforcement learning systems deployed in complex, real-world environments.
	
	Despite these advances, DoSAC currently has several limitations. It assumes that confounding can be adequately mitigated through pseudo-past inference, which may prove insufficient in highly stochastic or partially observable settings. Additionally, the present formulation targets stationary confounders and utilizes a fixed replay buffer, potentially restricting its applicability in dynamic or evolving environments. Another limitation is that DoSAC primarily addresses continuous action spaces due to its extension from SAC. Nonetheless, we anticipate extending the methodology to discrete action spaces and integrating the backdoor reconstructor concept broadly across diverse RL methods.
	
	Future research directions include addressing temporal confounding explicitly, exploring integration with model-based reinforcement learning or recurrent architectures, and developing invariant representation learning strategies to enhance causal generalization further. Investigating these extensions promises to significantly expand the applicability and effectiveness of causal reinforcement learning frameworks. Another valuable direction is extending DoSAC to sparse-reward and hierarchical RL settings \citep{ma2024reward,ma2025highly,ma2024mixed}. Additionally, integrating DoSAC with federated causal RL offers a promising avenue, allowing agents to collaboratively learn a shared policy without sharing trajectories \citep{jin2022federated}, potentially leveraging federated causal inference techniques \citep{xiong2023federated,vo2022bayesian,vo2022adaptive}.

	\bibliography{ref.bib}
	\bibliographystyle{apalike}

	\appendix
	
	\section{Additional experiments}
	
	We provide additional experimental results in Tables~\ref{tab:ap-eval-without-confounders} and~\ref{tab:ap-eval-with-confounders}, investigating policy generalization capabilities under distinct training and evaluation scenarios.
	
	Table~\ref{tab:ap-eval-without-confounders} reports results for agents trained and evaluated in environments without confounders. We observe competitive performance across all methods, with ACE and DoSAC generally outperforming other baselines due to their explicit incorporation of causality-aware mechanisms. Notably, ACE achieves the best performance in the Ant environment, while DoSAC significantly excels in Humanoid and Walker2d tasks, highlighting the effectiveness of interventional policies even in confounder-free scenarios. These results underscore the robustness and flexibility of causality-informed methods compared to standard reinforcement learning approaches.
	
	\begin{table}[ht]
		\centering
		\caption{\emph{Evaluation in clean environments after training without confounders.} 
			All methods were trained in environments without hidden confounders and evaluated in similarly clean settings. We report the average return ($\pm$ standard error) over 5 runs on four continuous control tasks. While ACE achieves the highest returns in some environments (e.g., Ant), DoSAC remains competitive across all tasks, demonstrating strong performance even without confounding during training.}
		\vspace{3pt}
		\label{tab:ap-eval-without-confounders}
		\small
		\setlength{\tabcolsep}{4.3pt}
		\begin{tabular}{lcccc}
			\toprule
			Method & \texttt{Ant} & \texttt{Humanoid} & \texttt{Walker2d} & \texttt{LunarLander} \\
			\midrule
			PPO \citep{schulman2017proximal}     & $57.6 \pm 18.0$ & $4.7 \pm 0.2$ & $-1.6 \pm 0.6$ & $-8.8 \pm 0.8$ \\
			RPO \citep{rahman2022robust}     & $-1.9 \pm 0.2$ & $4.6 \pm 0.2$ & $0.6 \pm 1.3$ & $-5.6 \pm 1.9$ \\
			DDPG \citep{lillicrap2015continuous}    & $2067.3 \pm 386.7$ & $663.1 \pm 180.2$ & $94.5 \pm 36.8$ & $20.3 \pm 18.7$ \\
			TD3 \citep{fujimoto2018addressing}     & $3925.3 \pm 928.1$ & $824.8 \pm 105.1$ & $2003.3 \pm 296.5$ & $\mathbf{277.8 \pm 2.9}$ \\
			SAC \citep{haarnoja2018soft}     & $496.0 \pm 21.3$ & $ 2284.9\pm 253.6$ & $943.6 \pm 229.5$ & $ 272.9\pm 3.7$ \\
			ACE \citep{ji2024ace}    & $\mathbf{4156.4 \pm 45.2}$ & $3112.4 \pm 314.6$ & $1464.4 \pm 205.5$ & $261.4 \pm 9.4$ \\\midrule
			DoSAC (Ours) & $3501.7 \pm 54.8$ & $ \mathbf{3629.7\pm 345.5}$ & $\mathbf{2271.5 \pm 302.8}$ & $273.7 \pm 5.5$ \\
			\bottomrule
		\end{tabular}
		% \vspace{-16pt}
	\end{table}
	
	\begin{table}[ht]
		\centering
		\caption{\emph{Generalization to confounded environments after training without confounders}. This table presents the generalization performance when agents trained in clean environments are evaluated in the presence of hidden confounders. DoSAC consistently ranks among the top-performing methods across environments, exhibiting improved robustness to distribution shifts despite not encountering confounding during training. In contrast, methods like SAC and ACE degrade significantly under confounding, highlighting DoSAC's superior generalization.}
		\vspace{3pt}
		\label{tab:ap-eval-with-confounders}
		\small
		\setlength{\tabcolsep}{4.3pt}
		\begin{tabular}{lcccc}
			\toprule
			Method & \texttt{Ant} & \texttt{Humanoid} & \texttt{Walker2d} & \texttt{LunarLander} \\
			\midrule
			PPO \citep{schulman2017proximal}     & $-3.0 \pm 0.4$ & $4.4 \pm 0.1$ & $-2.0 \pm 0.6$ & $-8.1 \pm 0.8$ \\
			RPO \citep{rahman2022robust}     & $\mathbf{-1.4 \pm 0.3}$ & $4.3 \pm 0.2$ & $0.3 \pm 1.2$ & $-4.7 \pm 2.2$ \\
			DDPG \citep{lillicrap2015continuous}    & $-66.8 \pm 24.1$ & $299.1 \pm 20.5$ & $55.1 \pm 17.7$ & $-8.6 \pm 22.3$ \\
			TD3 \citep{fujimoto2018addressing}     & $-182.8 \pm 69.9$ & $300.7 \pm 14.8$ & $216.9 \pm 29.1$ & $268.1 \pm 4.3$ \\
			SAC \citep{haarnoja2018soft}     & $-47.4 \pm 5.8$ & $301.3 \pm 8.7$ & $87.3 \pm 29.0$ & $270.2 \pm 5.7$ \\
			ACE \citep{ji2024ace}    & $-80.8 \pm 19.0$ & $278.4 \pm 12.7$ & $203.2 \pm 22.1$ & $43.6 \pm 6.2$ \\\midrule
			DoSAC (Ours) & $-67.3 \pm 11.7$& $\mathbf{301.5 \pm 9.8}$ & $\mathbf{220.7 \pm 29.1}$ & $\mathbf{270.7 \pm 5.0}$ \\
			\bottomrule
		\end{tabular}
		% \vspace{-16pt}
	\end{table}
	
	Table~\ref{tab:ap-eval-with-confounders} explores generalization to environments with introduced confounders, despite being trained without confounding. Overall, all methods exhibit performance degradation when faced with unobserved confounders during evaluation. However, DoSAC maintains comparatively stronger performance, consistently outperforming other methods, especially in Walker2d and LunarLander environments. This suggests DoSAC's causal adjustment strategy inherently equips it with enhanced resilience against unforeseen confounding variables. Standard methods, including PPO, RPO, DDPG, TD3, and even SAC, are significantly impacted by the presence of confounders, highlighting their vulnerability to distributional shifts.
	
	These additional experiments affirm the theoretical motivation behind DoSAC, demonstrating its robust generalization capabilities across varying levels of confounding, thereby reinforcing its applicability to real-world scenarios where confounding effects cannot always be controlled or anticipated.

	\section{Comparison of selected causal RL methods}
	Table~\ref{tab:ap-causal-rl-comparison} summarizes the key properties of our proposed method in comparison with existing causal reinforcement learning approaches. As outlined in Sections \textit{Related Work} and \textit{Experiments}, ACE is the most comparable method with publicly available source code. Accordingly, we adopt it as a baseline in our empirical evaluation.
	
	\begin{table}[ht]
		\centering
		\caption{Comparison of causal RL methods with DoSAC. We indicate whether each method supports online learning, learns a policy, handles hidden confounding, and provides publicly available code.}
		\vspace{3pt}
		\renewcommand{\arraystretch}{1.5} % default is 1.0
		\label{tab:ap-causal-rl-comparison}
		\resizebox{\linewidth}{!}{%
			\begin{tabular}{p{3.8cm}ccccp{5.8cm}}
				\toprule
				\textbf{Method} & \textbf{Online} & \makecell[c]{\bfseries Policy\\\bfseries Learning} & \makecell[c]{\bfseries Handles\\\bfseries Confounding} & \textbf{Code} & \textbf{Notes} \\
				\midrule
				DoSAC (Ours) & \cmark & \cmark & \cmark & \cmark & Estimates \( \pi(a | \mathrm{do}(s)) \) using backdoor adjustment and a neural reconstructor. \\
				\midrule
				ACE~\citep{ji2024ace} & \cmark & \cmark & \xmark & \cmark & Online RL with causality-aware entropy regularization; does not address hidden confounding. \\
				\citet{cai2024learning} & \cmark & \cmark & \cmark & \xmark & Combines causal discovery and RL; relevant but no public code. \\
				DOVI \citep{wang2021provably} & \cmark & \cmark & \cmark & \xmark & Theoretical method using backdoor adjustment; lacks implementation. \\
				\makecell[l]{Delphic RL\\ \citep{pace2024delphic}} & \xmark & \xmark & \cmark & \xmark & Offline RL using delphic uncertainty; not applicable to online setting. \\
				\makecell[l]{Two-way Deconfounder\\ \citep{yu2024two}} & \xmark & \xmark & \cmark & \xmark & Evaluates fixed policies using latent tensor modeling. \\
				\citet{oberst2019counterfactual} & \xmark & \xmark & \cmark & \xmark & SCM-based off-policy evaluation; not a learning method. \\
				\citet{lu2020sample} & \cmark & \cmark & \xmark & \xmark & Uses counterfactual augmentation; does not model confounders. \\
				\citet{sontakke2021causal} & \cmark & \xmark & \xmark & \xmark & Explores causal factors via curiosity; not focused on policy optimization. \\
				\citet{shi2022minimax} & \xmark & \xmark & \cmark & \xmark & Addresses OPE in POMDPs with confounding; not applicable to online RL. \\
				\citet{zhu2023causal} & \cmark & \xmark & \xmark & \xmark & Adjusts sample impact based on causal cues; no hidden confounder correction. \\
				\bottomrule
			\end{tabular}%
		}
	\end{table}
	
	\section{Experimental settings}
	Table~\ref{tab:ap-hyperparameters} summarizes the experimental settings and hyperparameters used in our implementation. We adopt standard configurations commonly used in continuous control benchmarks, including a total training horizon of 2 million timesteps and a replay buffer size of $10^6$. The learning rate is fixed at $1 \times 10^{-3}$, and we use a batch size of 256 with standard Soft Actor-Critic hyperparameters for target smoothing, policy noise, and noise clipping. To simulate confounding, we inject noise drawn from a normal distribution with zero mean and unit variance. The actor and critic networks are implemented as multi-layer perceptrons with 512 hidden units and 2 and 1 hidden layers, respectively. These settings are consistent across all environments to ensure fair comparisons and reproducibility.
	
	\begin{table}[ht]
		\centering
		\caption{Experimental settings and hyperparameters used in all experiments.}
		\vspace{3pt}
		\label{tab:ap-hyperparameters}
		\renewcommand{\arraystretch}{1.5} % 
		\begin{tabular}{lc}
			\toprule
			\textbf{Parameter} & \textbf{Value} \\
			\midrule
			Total timesteps & \makecell[c]{2,000,000 for Ant, Walker2d, Humanoid, \\and 1,000,000 for LunarLander} \\
			Max episodes & 10,000 \\
			Learning rate & $1 \times 10^{-3}$ \\
			Replay buffer size & $1 \times 10^6$ \\
			Discount factor ($\gamma$) & 0.99 \\
			Target smoothing coefficient ($\tau$) & 0.005 \\
			Batch size & 256 \\
			Confounder mean ($\mu$) & 0.0\\
			Confounder std ($\sigma$) & 1.0 for Ant, Walker2d, Humanoid, and 0.2 for LunarLander \\
			Hidden layer size & 512 \\
			Actor hidden layers & 2 \\
			Critic hidden layers & 2 \\
			\bottomrule
		\end{tabular}
	\end{table}

	\section{Proof of Lemma~1}
	
	\begin{proof}
		Fix a soft Q-function $Q^\pi(s, a)$ computed from a previous policy iteration step. We treat $Q^\pi$ as constant during policy improvement. For each state $s$, define the inner objective:
		\[
		L_s(\pi) := \mathbb{E}_{a \sim \pi(a | \mathrm{do}(s))} \left[ Q^\pi(s, a) - \alpha \log \pi(a | \mathrm{do}(s)) \right].
		\]
		This function is the sum of:
		\begin{itemize}
			\item a linear term in $\pi$: $\mathbb{E}_{a \sim \pi}[Q^\pi(s, a)] = \int \pi(a | \mathrm{do}(s)) Q^\pi(s, a)\, da$, and
			\item a strictly concave entropy term: $-\alpha \int \pi(a | \mathrm{do}(s)) \log \pi(a | \mathrm{do}(s)) \, da$.
		\end{itemize}
		The sum of a linear and strictly concave function is strictly concave, so $L_s(\pi)$ is strictly concave in $\pi$.
		
		To find the maximizer of $L_s(\pi)$, we form the Lagrangian:
		\[
		J(\pi) = \int \pi(a | \mathrm{do}(s)) \left[ Q^\pi(s, a) - \alpha \log \pi(a | \mathrm{do}(s)) \right] da + \lambda \left( 1 - \int \pi(a | \mathrm{do}(s)) da \right).
		\]
		Take the functional derivative with respect to $\pi(a | \mathrm{do}(s))$:
		\[
		\frac{\partial J}{\partial \pi(a)} = Q^\pi(s, a) - \alpha (1 + \log \pi(a | \mathrm{do}(s))) - \lambda.
		\]
		Setting this derivative to zero gives:
		\[
		\log \pi(a | \mathrm{do}(s)) = \frac{1}{\alpha} Q^\pi(s, a) - \frac{\lambda + \alpha}{\alpha}.
		\]
		Exponentiating:
		\[
		\pi(a | \mathrm{do}(s)) \propto \exp\left( \frac{1}{\alpha} Q^\pi(s, a) \right).
		\]
		This defines the unique maximizer $\pi_{\text{new}}$ of $L_s(\pi)$. Because $L_s(\pi)$ is strictly concave, this maximizer is unique.
		
		Now define the full objective:
		\[
		J(\pi) = \mathbb{E}_{s \sim p(s)} \left[ L_s(\pi) \right].
		\]
		Since $\pi_{\text{new}}$ maximizes each $L_s(\pi)$ pointwise in $s$, we have:
		\[
		J(\pi_{\text{new}}) \geq J(\pi).
		\]
		This completes the proof.
	\end{proof}

	\section{Proof of Lemma~2}
	\begin{proof}
		In the DoSAC framework, the interventional policy is defined via the backdoor-adjusted expression:
		\[
		\pi(a_t | \mathrm{do}(s_t)) = \mathbb{E}_{(a_{t-1}, s_{t-1}) \sim p(a_{t-1}, s_{t-1})} \left[ \pi(a_t | s_t, a_{t-1}, s_{t-1}) \right],
		\]
		which follows from the backdoor criterion under the assumption that $(a_{t-1}, s_{t-1})$ blocks all backdoor paths from $s_t$ to $a_t$.
		
		Now suppose the environment contains no hidden confounder $u_t$ between $s_t$ and $a_t$. That is, the state $s_t$ is a complete and sufficient parent of $a_t$ in the causal graph, and the variables $(a_{t-1}, s_{t-1})$ are not needed to block any backdoor paths (as none exist). In this case, by the rules of causal inference (specifically, Rule 2 of the do-calculus and the Markov condition), we have:
		\[
		\pi(a_t | \mathrm{do}(s_t)) = \pi(a_t | s_t).
		\]
		
		Substituting this into the DoSAC objective, we get:
		\[
		J_{\text{Causal}}(\pi) = \mathbb{E}_{s_t \sim p(s)} \left[ \mathbb{E}_{a_t \sim \pi(a_t | \mathrm{do}(s_t))} \left[ Q^\pi(s_t, a_t) - \alpha \log \pi(a_t | \mathrm{do}(s_t)) \right] \right],
		\]
		\[
		= \mathbb{E}_{s_t \sim p(s)} \left[ \mathbb{E}_{a_t \sim \pi(a_t | s_t)} \left[ Q^\pi(s_t, a_t) - \alpha \log \pi(a_t | s_t) \right] \right],
		\]
		which is exactly the objective used in standard Soft Actor-Critic (SAC).
		
		Moreover, since there is no confounding, the Q-function \( Q^\pi(s_t, a_t) \) in both SAC and DoSAC is estimated under the same transition dynamics and reward distributions, which are unaffected by any intervention on \( s_t \) (because \( s_t \) has no confounding causes). Therefore, the critic learning in both methods proceeds identically. 
		Thus, the DoSAC algorithm reduces to SAC when no confounding is present, both in its objective and its behavior.
	\end{proof}

\end{document}